\def\year{2022}\relax
\documentclass[letterpaper]{article} 
\usepackage{aaai22}  
\usepackage{times}  
\usepackage{helvet}  
\usepackage{courier}  
\usepackage[hyphens]{url}  
\usepackage{graphicx} 
\def\UrlFont{\rm}  
\usepackage{natbib}  
\usepackage{caption} 
\DeclareCaptionStyle{ruled}{labelfont=normalfont,labelsep=colon,strut=off} 
\usepackage[utf8]{inputenc} 
\usepackage[T1]{fontenc}    
\usepackage{hyperref}       
\usepackage{url}            
\usepackage{booktabs}       
\usepackage{amsfonts}       
\usepackage{nicefrac}       
\usepackage{microtype}      
\usepackage{xcolor}         
\usepackage{algorithm,algpseudocode}
\usepackage{graphicx}
\usepackage{booktabs} 

\usepackage{multirow}
\RequirePackage{amsmath,dsfont}
\usepackage{amsthm}
\usepackage{amssymb,accents}
\usepackage{tikz}
\usepackage{paralist}
\usepackage{optidef}
\usepackage{wrapfig}

\usepackage{caption}

\usepackage{subfig}
\usepackage{xcolor}
\usepackage{colortbl}
\usepackage{paralist}
\usepackage{multirow}
\usepackage{graphicx,verbatim,enumitem}
\usepackage{amsthm, amsmath, amsfonts, amssymb}
\usepackage{caption}
\usepackage[utf8]{inputenc} 
\usepackage[T1]{fontenc}    
\usepackage{url}            
\usepackage{booktabs}       
\usepackage{nicefrac}       
\usepackage{microtype}      
\usepackage{algorithm,algorithmicx,algpseudocode}

\usepackage{caption, tabularx}
\usepackage{enumitem}

\usepackage{ulem} 

\algdef{SE}[DOWHILE]{Do}{doWhile}{\algorithmicdo}[1]{\algorithmicwhile\ #1}%

\input{definitions.tex}

\def\C{{\mathcal{C}}}
\def\xhat{\hat{x}}
\def\ecl{\textit{Ecl~}}
\def\dlip{\textit{DuaLip~}}
\def\scs{\textit{SCS~}}

\def\Ctil{\tilde{C}}
\def\xtil{\tilde{x}}

\def\K{\mathcal{K}}
\def\ktil{\tilde{k}}

\def\Box{{\it Box~}}
\def\SE{{\it Simplex-Eq~}}
\def\SI{{\it Simplex-Iq~}}
\def\BE{{\it Box-Cut-Eq~}}
\def\BI{{\it Box-Cut-Iq~}}
\def\Ass{{\it Assignment~}}
\def\Slate{{\it Slate~}}
\def\GP{{\it General Polytope~}}

\title{Efficient Vertex-Oriented Polytopic Projection for Web-scale Applications}

\author {
    Rohan Ramanath\thanks{Work done while at LinkedIn },
    S. Sathiya Keerthi,
    Yao Pan,
    Konstantin Salomatin,
    Kinjal Basu
}
\affiliations {
    LinkedIn Corportation\\
    Mountain View, CA, USA\\
    rohan@ramanath.me, 
    \{keselvaraj, yopan, ksalomatin, kbasu\}@linkedin.com
}

\usepackage{bibentry}

\begin{document}

\maketitle

\begin{abstract}

We consider applications involving a large set of instances of projecting points to polytopes. We develop an intuition guided by theoretical and empirical analysis to show that when these instances follow certain structures, a large majority of the projections lie on vertices of the polytopes. To do these projections efficiently we derive a vertex-oriented incremental algorithm to project a point onto any arbitrary polytope, as well as give specific algorithms to cater to simplex projection and polytopes where the unit box is cut by planes. Such settings are especially useful in web-scale applications such as optimal matching or allocation problems. Several such problems in internet marketplaces (e-commerce, ride-sharing, food delivery, professional services, advertising, etc.), can be formulated as Linear Programs (LP) with such polytope constraints that require a projection step in the overall optimization process. We show that in the very recent work \cite{basu2020}, the polytopic projection is the most expensive step and our efficient projection algorithms help in gaining massive improvements in performance.
\end{abstract}

\setlength{\abovedisplayskip}{4pt}
\setlength{\belowdisplayskip}{4pt}

\section{Introduction}
\label{sec:introduction}
\def\xtilde{\tilde{x}}
Euclidean projection onto a polytope is not only a fundamental problem in computational geometry \cite{he2006projection, rambau1996projections}, but also has several real-life applications ranging from decoding of low-density parity-check codes \cite{wei2017iterative} to generating screening rules for Lasso \cite{wang2013lasso}. Recent work \cite{basu2020, boyd-admm} also uses polytope projection as an operation in solving large optimization problems. Several such large-scale optimization solvers which use polytopic projection \cite{basu2020, boyd-admm, parikh2014block} are being used to tackle different problems in the internet industry such as matching \cite{azevedo2016matching, makari2013distributed} and multi-objective optimization \cite{agarwal2011click, agarwal2012personalized}. Other examples include problems in natural language processing like structured prediction \cite{smith:2011:synthesis, rush:2012:tutorial,martins:2015:ad3}, semi-supervised approaches \cite{chang:2013:tractable}, multi-class/hierarchical classification~\cite{selvaraj:2012:extension}, etc.

Given a compact set $\Ccal$ and a point $x$, let $\Pi_\Ccal(x)$ denote the (euclidean) projection of $x$ onto $\Ccal$, i.e., $\Pi_\Ccal(x) = \arg\min_{y\in \Ccal} \|y-x\|$. Certain large scale optimization problems, especially those arising from applications in the internet industry, have a huge number of polytope constraint sets, $\{\Ccal_i\}$. To solve such problems each iteration of an algorithm requires the projection of points $\{x_i\}$ to $\{ \xtilde_i = \Pi_{\Ccal_i}(x_i)\}$. Computing such sets of projections is usually the key bottleneck~\cite{wei2017iterative} and hence efficiently solving the projections is crucially important to solve such large-scale optimization problems.

The set of points, $\{ x_i \}$ usually has some structure connecting the $x_i$ that dictates the nature of the projections. Most general purpose solvers such as ADMM \cite{boyd-admm}, Block Splitting \cite{parikh2014block} and Splitting Conic Solver \cite{scs} do not exploit this special structure. Very recently, Basu et. al. \cite{basu2020} proposed ECLIPSE (shorthand \textit{Ecl}) - to solve such large scale problems while exploiting some general overall structure, but still used off-the-shelf projection solvers such as the one in \cite{duchi2008efficient} to tackle the projection step. As a result, similar to the general purpose algorithms \cite{boyd-admm, scs, parikh2014block}, the major computational bottleneck for \ecl is also the projection step.

Although there are many different forms of projection structures, in this paper we focus on a particular structure motivated by applications in the internet industry. We consider those, for which, a majority of the projections, $\xtilde_i$ lies on a vertex of $X_i$ and also, over all $i$, the mean dimension of the minimal face~\cite{grunbaum2003convex} of $X_i$ that contains $\xtilde_i$ is very small. 
We refer to this as the {\it vertex oriented structure}. Even for special polytopes such as the simplex  and {\it box-cut} (a hypercube cut by a single plane) general purpose projection algorithms~\cite{duchi2008efficient} are not designed to be efficient for the above situation. The main goal of this paper is to develop  special purpose projection algorithms that (a) are specially efficient for the vertex oriented structure; and (b) are applicable to special polytopes such as the simplex, {\it box-cut}, and general polytope forms.

To show the efficacy of these special purpose projection algorithms, we switch out the projection step of \ecl, and run an ablation study over various internet marketplace problems. We call this new solver \textit{DuaLip}, a \textbf{Dua}l decomposition-based \textbf{Li}near \textbf{P}rogramming framework which shows drastic speedups over the existing state-of-the-art. Moreover, \ecl only tackled simple polytope constraints such as the unit box or the simplex, which not only reduced the generality, but also made the solver tough to use right out of the box. By 
replacing the projection component with the novel algorithms developed in this paper,
\textit{DuaLip} is now capable of solving a much wider range real-world applications, much more efficiently.

The rest of the paper is organized as follows;  We first introduce the problem, the motivating applications, and describe the \ecl algorithm. We then focus on the efficient projection algorithms in detail and specialize the general solution to special structured polytopes. We empirically show the impact of these algorithms on web-scale data, before finally concluding with a discussion. All formal proofs are given in the full paper \cite{dualip}.

\definecolor{Gray}{gray}{0.9}
\section{Problem Setup}
\label{sec:problem}
We begin with some notation. Let $x \in \real^K$ denote the point that we wish to project onto a compact polytope $\Ccal$. The projection operator $\Pi_\Ccal(x)$ can be written as $\Pi_\Ccal(x) = \argmin_{y \in \Ccal} \| y - x \|$
where $\|\cdot\|$ denotes the Euclidean distance. Throughout this paper, we consider the following set of polytopes: 
\begin{compactenum}
\item \Box:  $\C = \{ x: 0 \le x_k \le 1 \; \forall k \}$ 
\item \SE: $\C = \{ x: x_k \ge 0 \; \forall k, \; \sum_k x_k = 1 \}$ 
\item \SI: $\C = \{ x: x_k \ge 0 \; \forall k, \; \sum_k x_k \le 1 \}$ 
\item \BE: $\C = \{ x: 0 \le x_k \le 1 \; \forall k, \; \sum_k x_k = \delta \}$ ($\delta=$ positive integer, $1<\delta<K$)
\item \BI: $\C = \{ x: 0 \le x_k \le 1 \; \forall k, \; \sum_k x_k \le \delta \}$ ($\delta=$ positive integer, $1<\delta<K$) 
\item \textit{Parity Polytope}: $\C = co \, \{v_r\}$ where the $v_r$ are binary vectors with an even number of 1s and $co$ denotes convex hull.
\item \GP: $\C = co \, \{ v : v\in V \}$ where $V$ is a finite set of polytope vertices. 
\end{compactenum}
Here we use $E,I$ to denote an equality or inequality sign denoting whether we are interested in the surface or the closed interior of the polytope. Such polytopes naturally occur as constraints in different optimization formulations in the internet industry. We motivate the need of such polytopes through typical recommender system problems. 

We denote users by $i = 1,\ldots,I$ and items by $k = 1, \ldots, K$. Let $x_{ik}$ denote any association between user $i$ and item $k$, and be the variable of interest. For example, $x_{ik}$ can be the probability of displaying item $k$ to user $i$. The vectorized version is denoted by $x = (x_1, \ldots, x_I)$ where $x_i = \{x_{ik}\}_{k=1}^{K}$. Throughout this paper we consider problems of the form:
\begin{equation}
\label{eq:main_problem}
\begin{aligned}
\min_{x}~c^T x~~~\text{s.t.}~~~Ax \leq b,~~x_i \in \mathcal{C}_i,~i\in [I],
\end{aligned}
\end{equation}
where, $A_{m \times n}$ is the constraint matrix, $b_{m \times 1}$ and $c_{n \times 1}$ are the constraint and objective vectors respectively, and ${\mathcal C}_{i}$ are compact polytopes. $x \in \real^n$ is the vector of optimization variables, where $n=IK$ and $[I] = \{1, \ldots, I\}$.
\subsection{Applications}
\label{subsec:applications}

\begin{table*}[!h]
\vspace{-0.2cm}
\centering
\begin{tabular}{p{6cm}|p{2.2cm}|p{2.4cm}|c|c|c} 
\hline
\textbf{Application} & \textbf{Objective} ($c_{ik}$)  & \textbf{Constraint} ($a_{ik}$) & \textbf{Projection} & $\boldsymbol{K}$ & $\boldsymbol{n}$ \\
\hline
Email Optimization \cite{basu2020} & sessions & unsubscribes & \Box  & $100$ & $10b$ \\
Diversity in Network Growth & connection  & invitation & \BE  & $20k$ & $2t$ \\
Item Matching (Jobs Recommendation) & job applies & budget & \BI &  $1m$ & $100t$ \\
Multiple Slot Assignment (Feed) & engagement & revenue & \Ass  & $10b$ & $1q$ \\
Slate Optimization (Ads Ranking) & revenue &  budget & \Slate &  $10b$ & $1q$ \\
\hline
\end{tabular}
\caption{ \small {A list of applications with constraints, projections and size. Here we use \textit{-Eq,-Iq} to denote an equality or inequality sign in the constraints (see Section \ref{sec:projection}). Finally the short hand, $k,m,b,t,q$ denotes thousand, million, billion, trillion and quintillion ($10^{18}$), respectively.}}
\label{tab:applications}
\vspace{-0.4cm}
\end{table*} 

Basu et. al. \cite{basu2020} described two major classes of problems, volume optimization (focused on email/notifications) and optimal matching (for network growth). We cover a much larger class of problems using a variety of different polytopic constraints.

\textbf{Diversity in Network Growth:} ``Rich getting richer'' is a common phenomenon in building networks of users \cite{fleder2009blockbuster}. Frequent or power users tend to have a large network and get the most attention, while infrequent members tend to lose out. To prevent these from happening, and to improve the diversity in the recommended users, we can frame the problem as a LP: 
\begin{subequations}
\label{eq:pymk}
\begin{align*}
\max_{x} \quad & \sum\nolimits_{ik} x_{ik} c_{ik}  \qquad\qquad\;\;\;\;\; \text{(Total Utility)}  \\
\text{s.t.} \quad &\sum\nolimits_i x_{ik} a_{ik} \geq b_k ~ \forall k \qquad \text{(Infrequent User Bound)}\\
 \quad &\sum\nolimits_{k} x_{ik} = \delta_i, \;\; 0 \leq x_{ik} \leq 1 
\end{align*}
\end{subequations}
where $c,a$ are utility and invitation models, $b_k$ denotes the minimum number of invitations to the $k$-th infrequent user, and $\delta_i$ denotes the number of recommendations to the $i$-th user. This makes $\mathcal{C}_i$ the  \BE polytope.

\textbf{Item Matching in Marketplace Setting:} In many two-sided marketplaces, there are creators and consumers. 
Each item created has an associated budget and the problem is to maximize the utility under budget constraints. For example, in the ads marketplace, each ad campaign has a budget and we need to distribute impressions appropriately. For the jobs marketplace, each paid job has a budget and the impressions need to be appropriately allocated to maximize job applications. 
Each of these problems can be written as 
\begin{align}
\label{eq:marketplace}
\max_{x} \;\; & \sum\nolimits_{ik} x_{ik} c_{ik} \;\; \nonumber\\
\text{s.t.} \;\; & \sum\nolimits_{i} x_{ik} a_{ik} \leq b_k \; \forall k\in [K], \\
& \sum\nolimits_{k} x_{ik} \leq \delta_i \text{ and } 0 \leq x_{ik} \leq 1 \nonumber
\end{align}
where $c,a$ are models estimating utility and budget revenue, respectively and $\delta_i$ is the maximum number of eligible items to be shown to the $i$-th member. Here the polytope constraint is \BI.

\textbf{Multiple Slot Assignment:} 
This is an extension of the item ranking problem where the utility of an item depends on the position in which it was shown \cite{keerthi2007}. Consider for each request or query $i$, we have to rank $L_i$ items ($\ell = 1, \ldots, L_i$) in $S_i$ slots ($s = 1,\ldots, S_i$). We want to maximize the associated utility subject to constraints on the items. Mathematically this bipartite structure of multiple items and slots can be framed as:   
\begin{subequations}
\label{eq:assignment}
\begin{align*}
\max_{x} \;\; & \sum\nolimits_{i\ell s} x_{i\ell s} c_{i \ell s}  \\
\text{s.t.} \;\; & \sum\nolimits_{i \ell s} x_{i\ell s} a_{i\ell s}^{(j)} \leq c_j ~~~~~  \forall j\in [J]\\
& \sum\nolimits_{s} x_{i\ell s} = 1, \sum\nolimits_{\ell} x_{i\ell s} = 1 \text{ and } x_{i\ell s} \geq 0 \; \forall i,\ell,s
\end{align*}
\end{subequations}
where $c$ and $a^{(j)}$ are the associated utility and $j$-th constraint values. Note that the projection set $\mathcal{C}_i$ captures the need to show each item and the fact that each slot can contain only one item. 

This formulation involves a special $\C_i$ and hence, special projections. We avoid this by using a revised formulation that introduces an index $k = 1, \ldots, K$ with each $k$ denoting a set of distinct items (instead of being a single item) in the set of slots. The optimal assignment is then choosing one such assignment per request. Thus we have,
\begin{align}
\label{eq:simplex-slate}
&\max_{x} & & \sum\nolimits_{ik} x_{ik} c_{ik}  \nonumber \\
& \text{s.t.} & & \sum\nolimits_{ik} x_{ik} a_{ik}^{(j)} \leq c_j ~~~~~  \forall j\in [J] \nonumber \\
& & & \sum\nolimits_{k} x_{ik} = 1 \text{ and } x_{ik} \geq 0 \; \forall i,k
\end{align}

\textbf{Slate Optimization:} This is a variant of the multiple slot assignment problem, where each of the $L_i$ items are ranked and each index $k$ (slate) corresponds to a ranked selection of distinct items to be placed in the slots. This can be set up as a path search on a DAG; see \cite{keerthi2007} for more details. Both the Slate and Assignment problems can be written using the \SE polytopic constraint. Table \ref{tab:applications} describes these applications with respect to the associated models, projection, and problem size. 

In the subsequent sections, we describe how we develop new projection algorithms to handle such larger classes of problems. To demonstrate the value of these novel algorithms, we mainly work with a proprietary \textbf{Jobs Matching Dataset} ($D$). Here, we are trying to solve the problem of the form in \eqref{eq:marketplace}, where instead of the \textit{Box-Cut} constraint we consider the simplex constraint: $\sum\nolimits_{k} x_{ik} = 1 \text{ and } 0 \leq x_{ik} \leq 1$ and we have $n = 100$ trillion. We rely on a real-world dataset only to represent the scale of the problem, i.e. the number of coupling and non-coupling constraints seen in practice. The data does not contain any personally identifiable information, respects GDPR and does not contain any other user information. To promote reproducible research we open source (with FreeBSD License)\footnote{ https://github.com/linkedin/dualip} the solver with all the efficient projection operators discussed in the paper. There, we report the solver's performance on the open-source movie-lens dataset \cite{harper2015movielens}, which contains user rating of movies to formulate an optimization problem (taking a similar approach to \cite{makari2013distributed}). Movie-lens is a public domain matching problem of a much smaller scale than seen in internet marketplaces.

\subsection{Solving the LP}
\label{subsec:solution}
For sake of completeness, we briefly describe \ecl algorithm here. To solve problem \eqref{eq:main_problem}, \ecl introduces a new perturbed problem
\begin{equation}
\label{eq:qp}
\min_{x}~~c^T x + \frac{\gamma}{2} x^Tx~~\text{s.t.}~~Ax \leq b,~x_i \in \mathcal{C}_i, i \in [I]
\end{equation}
where $\gamma > 0$ helps to make the dual objective function smooth; $\gamma$ is kept small to keep the solution close to that of the original problem \eqref{eq:main_problem}.
To make the problem \eqref{eq:qp} amenable to first order methods, \ecl considers the Lagrangian dual,
\begin{align}
\label{eq:dualfunc}
g_{\gamma}(\lambda) = \min_{x \in \mathcal C} ~~ \left\{ c^T x + \frac{\gamma}{2} x^Tx + \lambda^T(Ax-b) \right\},
\end{align}
where $\mathcal{C} = \Pi_{i=1}^I \mathcal{C}_i$. Now, by strong duality~\cite{boyd2004convex}, the optimum objective $g_{\gamma}^*$ of the dual
\begin{align}
    \label{eq:dual}
    g_{\gamma}^*:=\max_{\lambda \geq 0} ~ g_{\gamma}(\lambda)
\end{align} 
is the minimum of~\eqref{eq:qp}. \ecl shows that $\lambda \mapsto g_{\gamma}(\lambda)$ is differentiable and the gradient is Lipschitz continuous. Moreover, by Danskin's Theorem \cite{danskin2012theory} the gradient can be explicitly expressed as,
$\nabla g_{\gamma}(\lambda) = Ax_\gamma^*(\lambda) -b$
 where,
\begin{align}
x_{\gamma}^*(\lambda) &= \argmin_{x \in \mathcal C} ~~ \left\{ c^T x + \frac{\gamma}{2} x^Tx + \lambda^T(Ax-b) \right\} \nonumber\\
&= \big\{
\Pi_{\mathcal{C}_i}[-\tfrac{1}{\gamma}(A_i^T\lambda + c_i)]
\big\}_{i=1}^I
\label{eq:xstar}
\end{align} 
where $\Pi_{\mathcal{C}_i}(\cdot)$ is the Euclidean projection operator onto  $\mathcal{C}_i$, and, $A_i$, $c_i$ are the parts of $A$ and $c$ corresponding to $x_i$. Based on this \ecl uses accelerated gradient method as the main optimizer to solve the problem in \eqref{eq:dual}. 
For more details, we refer to \cite{basu2020}.


\subsection{Cost Profiling}
On large datasets such as $D$ we find that \ecl takes days to solve.
This motivates us to understand which parts can be improved to get the biggest overall speedup.
We profile each of the parts 
to analyze the performance bottlenecks of \ecl. 
Table~\ref{tab:complexity} shows the complexity of each term required to compute the gradient along with the sample time spent per iteration of the solver. It is clear that the optimizer's internal update steps only take $~3\%$ of the time while the rest ($97\%$) is used to compute the gradient (columns 2 to 6 in Table~\ref{tab:complexity}). The results are consistent on other datasets as well when the number of executors is tuned appropriately.
\newcolumntype{g}{>{\columncolor{Gray}}c}
\begin{table*}[!th]
\centering
\begin{tabular}{|l|c|c|c|c|c|g|}
\hline
Profiling & $A_i^T\lambda$ & $\hat{x}_i(\lambda)$ & $A_i\hat{x}_i$ & $A\hat{x}, c\hat{x}, ||\hat{x}||$& $\lambda^T (Ax-b)$ & LBFGS \\
\hline
Complexity & $\mathcal{O}(k)$ & $\mathcal{O}(k \log k)$ & $\mathcal{O}(\mu_i)$ & $\mathcal{O}(I/w) + \mathcal{O}(w)$ & $\mathcal{O}(K)$ &$\mathcal{O}(K)$\\
Sample time & 10\% & 74\% & 2\% & 8\% & 3\% & 3\% \\
\hline
\end{tabular}
\caption{\small{Complexity analysis of different components on dataset $D_1$ with $w=100$ nodes. $k$, $\mu_i$ as defined later. (a) In gray is the time spent inside the optimizer, (b) in white is the time taken to compute the gradient. Notation as defined in ($\ref{proj:eq1}$)}}  
\label{tab:complexity}
\vspace{-0.2cm}
\end{table*}
For the rest of paper, we only focus on the most expensive ($74\%$) step in the gradient computation, that is the projection operation to compute $x_\gamma^*(\lambda)$ as in \eqref{eq:xstar}.

\section{Efficient Projection Algorithms}
\label{sec:projection}
Recall the projection problem to be solved: for each $i$, we want to find
\begin{equation}
    (x_\gamma^*(\lambda))_i = 
\Pi_{\mathcal{C}_i}[-\tfrac{1}{\gamma}(A_i^T\lambda + c_i)]
 = \arg\min_{x_i\in \C_i} \| x_i - \xhat_i \|^2
    \label{proj:eq1}
\end{equation}
where $\xhat_i = -\frac{1}{\gamma}(A_i^T\lambda + c_i)$.
To simplify notation, we will leave out the `$i$' unless it is necessary. 
The cost profiling shows that the projection step forms the bulk of the overall algorithm cost and hence, improving its speed would lead to overall algorithmic speedup. If projections form $\phi$ fraction of the total solution cost, then by making projections extremely efficient one can hope to get up to a speedup of $1/(1-\phi)$. We now focus on designing efficient algorithms that are well suited to marketplace problems. 
\ecl covers \Box and \SE. The \Box projection solution is very efficient, 
but in our case the \SE projection algorithm \cite{duchi2008efficient} is not. 
This is primarily because, for a large fraction of $i$'s, the projections are at a vertex of $\mathcal{C}_i$, which can be identified in a more efficient way. 






{\bf The Main Intuition:} 
We define a {\it corral}~\cite{chakrabarty2014provable} as the convex hull of a subset of the vertices of $\C_i$ with least cardinality that contains the projection of $\xhat_i$. Thus, if the projection is a vertex, then the vertex itself is the corral; if the projection lies in the relative interior of an edge formed by two vertices, then the corral is that edge, and so on.
The dimension of the corral is one less than the cardinality of its vertex subset. A vertex is a corral of dimension zero, an edge is a corral of dimension 1, and so on. Clearly there exists a corral that contains the projection, $x_i^*$. 

\def\dtil{\tilde{d}}

{\it Our efficient approach is based on the intuition that, for small $\gamma$, the mean dimension of the corral over all $i$ is small.} (As already mentioned below (\ref{eq:qp}), the value of $\gamma$ is chosen to be close to zero.) Since this is the main basis of the paper, we explain the reasoning behind the intuition. We begin by giving a simple example in 2d. Then we state a formal result that says what happens at $\gamma=0$. We follow that by taking the case of \SE and demonstrating how, as $\gamma$ becomes larger than zero, the mean corral dimension only increases gradually with $\gamma$. Further, we give strong empirical support for the intuition. After that, the rest of the section and \S\ref{sec:structured} focus on developing efficient projection algorithms based on the intuition.

\begin{figure}[!ht]
\begin{center}
\centerline{\includegraphics[width=\columnwidth,trim={0 3.7cm 1.8cm 0}, clip]{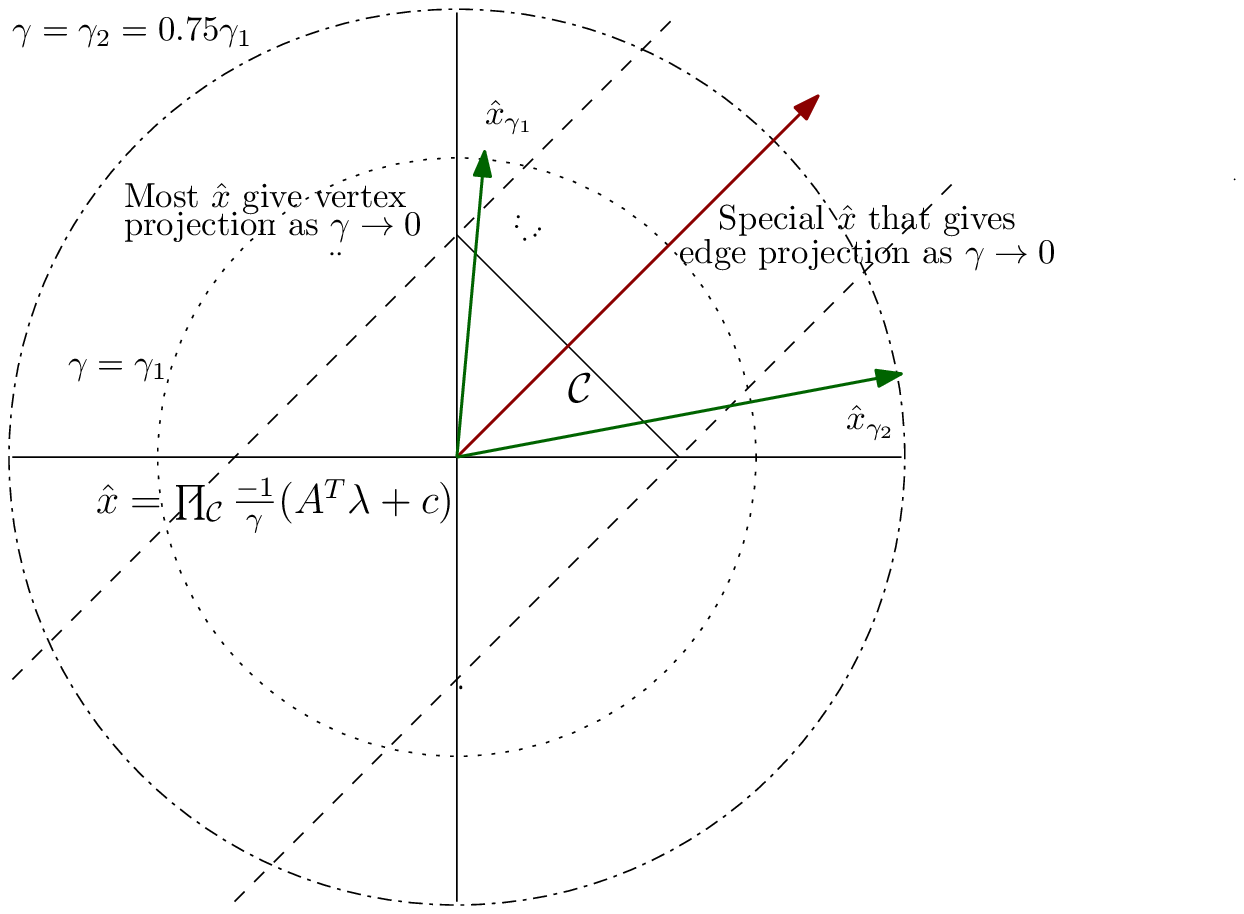}}
\caption{For a fixed feasible $\lambda$, the set of possible placements of $\xhat$ (see (\ref{proj:eq1}) and below it for the definition of $\hat{x}$) that project to the relative interior of the simplex edge (corral of dimension = 1) is only the infinite rectangle (shown using dashed lines) perpendicular to the edge. Unless $\xhat$ is exactly perpendicular to the edge (shown in red), when $\gamma$ becomes small, $\xhat$ crosses out of that infinite rectangle, after which a vertex becomes the projection. }
\label{twodExample}
\end{center}
\vspace{-0.5cm}
\end{figure}

Let us begin by illustrating the intuition using an example of \SE with $K=2$; see Figure~\ref{twodExample}. From the figure it is clear that, for small values of $\gamma$, the probability that the projection lies on a vertex (corral of dimension zero) is high. 

We now state a formal result for the $\gamma=0$ case.
The proof of Theorem 1 is given in the full paper \cite{dualip}.

\begin{theorem}
\label{thm:corral}
Let: $x \in \mathcal{C}$ be expressed as $Dx \le d$, $Ex = e$, $p=(b, c, d, e)$, $x^*(p)$ denote the optimal solution of \eqref{eq:main_problem}, and $x_0^*(p,\lambda) = \arg\min_{x\in\C} (c+A^T\lambda)^Tx$ (also same as (\ref{eq:xstar}) with $\gamma\rightarrow 0$).
For any $x$ define $\mu(x) = \sum_i\mu_i/I$ to be the mean corral dimension where $\mu_i$ is the corral dimension of $x_i$. 
Let $B_{\epsilon}(z)$ denote the open ball with center $z$ and radius $\epsilon$.
Given $\epsilon_b, \epsilon_c, \epsilon_d, \epsilon_e > 0$, if we define 
$\Bcal := \{ (\btil,\ctil,\dtil,\etil): \btil \in B_{\epsilon_b}(b), \ctil \in B_{\epsilon_c}(c), \dtil \in B_{\epsilon_d}(d), \etil \in B_{\epsilon_e}(e) \},$ then
\begin{compactenum}
\item $x^*(p)$ is unique and $\mu(x^*(p)) \le m/I$ for almost all\footnote{`almost all' is same as `except for a set of measure zero'.} $p \in \Bcal$.
\item  $x_0^*(p,\lambda)$ is unique and $\mu(x_0^*(p,\lambda)) = 0$, for almost all $p \in \Bcal$ and $\lambda\in \real^m$.
\end{compactenum}
\end{theorem}

\def\Fcal{{\cal{F}}}

Note that since each $\mu_i \in \mathbb{Z}$, 
it directly follows from Theorem \ref{thm:corral} that $\Itil$, the number of $i$'s with vertex solutions ($\mu_i = 0$) is at least $I-m$.
Moreover, $K$, which denotes the dimension of the $x_i$, does not affect $\mu$ at all (due to the structure of $\Ccal_i$ and the repeating nature in $i$). This is beneficial since $K$ is very large in \Ass and \Slate projections (see Table \ref{tab:applications}). 
Comparing results 1 and 2 of Theorem \ref{thm:corral}, we can see that non-vertex solutions start appearing as the solution approaches $\lambda_0 = \arg\max_{\lambda\ge 0} g_0(\lambda)$, and even there, it is well bounded. In  web-scale marketplace applications $m/I$ is very small, say, smaller than $0.01$. Thus, vertex projections dominate and they occur in more than 99\% of the $i$'s even as we come close to $\lambda_0$. 

A limitation of the theorem is that it gives a bound on $\mu$ only at $\gamma = 0$.  On the other hand, in our solution process, we use small $\gamma > 0$.
By Lemma 2 of \cite{basu2020} (also see \cite{mangasarian1979nonlinear, friedlander2008exact}), the solution $x^*$ of \eqref{eq:main_problem} remains unchanged under the $\gamma$ perturbation for sufficiently small positive $\gamma$ values (Figure~\ref{twodExample} gives the rough intuition), and so part 1 of the theorem applies for such positive $\gamma$.

Even the above formal result is limited. To better explain what happens as we gradually increase $\gamma$ above zero, let us analyze the \SE case.
Let us take any one $i$, but omit the subscript $i$ to avoid clumsiness. Let $z=-(A^T\lambda + c)$ so that $\hat{x}=z/\gamma$. We invoke Algorithm~\ref{alg:duchimod} (given in the next section) to give a precise analysis. Given a $\lambda$ and hence, given $\hat{x}$, we can use Algorithm~\ref{alg:duchimod} to quantify when a vertex (corral of dimension 0) will be the projection, when an edge (corral of dimension 1) will contain the projection, etc. In general position (think of random infinitesimal perturbation of $c$)  there will be a positive separation between the components of $z$. Without loss of generality let us take the components of $z$ to be ordered: $z_1 > z_2 > z_3 > \ldots$. With this in place, it is easy to use Algorithm~\ref{alg:duchimod} and derive the following: (a) Projection will be a vertex if $0 < \gamma \le (z_1 - z_2)$; (b) Projection will be on an edge if $(z_1 - z_2) < \gamma \le (z_1 - z_3) + (z_2 - z_3)$; (c) and so on. Thus projections start lying on higher dimensional corrals only slowly as $\gamma$ is increased. There is a graded monotonic increase in the optimal corral dimension with respect to $\gamma$.

It is possible to give similar details for constraint sets other than \SE; it is just that they get quite clumsy. In addition to what is given above, the other key point is that, as we move towards optimality, part 1 of Theorem 1 says that only a few $i$'s (at most $m$ of them at optimality where $m$ is the number of constraints in $Ax\le b$) will have non-vertex projections. Going by the graded effect of positive $\gamma$ on optimal corral dimensions, we get useful indications of what small values of $\mu$ we can expect as we close in on optimality with small (not necessarily extremely small) $\gamma$ (also see the empirical results in Table~\ref{tab:corral}).

Let us use empirical analysis to further understand what happens for larger $\gamma$ perturbations. Table~\ref{tab:corral} studies the behavior of $\mu$ along optimization paths of $\max_\lambda g_\gamma(\lambda)$ for a range of $\gamma$ values, on the representative dataset $D$. Even with large $\gamma$ values, note that $\mu$ is usually small (less than $3$).  Obtaining a theoretical bound on $\mu$ as a function of $\gamma$ is an interesting topic of future academic research. Such theory can be useful in many optimization formulations that invoke projection as a substep.

\vspace{-0.2cm}
\begin{table}[ht]
    \centering
    \begin{tabular}{|l|c|c|c|} 
\hline
 Corral metric & $\gamma_1=1$ & $\gamma_2=0.1$ & $\gamma_3=0.01$ \\
\hline
 $\mu$ & 2.55 & 0.12 & 0.03  \\
 $\Itil / I$ & 0.4522 & 0.9232 & 0.9676 \\
\hline
\end{tabular}
    \caption{\small {Behavior of the mean corral dimension, $\mu$ along optimization paths of $g_\gamma$ starting from $\lambda=0$ for three values of $\gamma$ s.t. $\gamma_1 > \gamma_2 > \gamma_3$. 
    }}
\label{tab:corral}
\vspace{-0.4cm}
\end{table}

\subsection{Vertex-Oriented Approach} 
Guided by the theory and the empirical observations that (a) vertex solutions occur most frequently and (b) the mean corral size is small, we devise the powerful approach to projection described in Algorithm~\ref{alg:ourapp}.

    \begin{algorithm}[H]
       \caption{\dlip vertex-first projection}
       \label{alg:ourapp}
    \begin{algorithmic}[1]
        \State Let $v^0$ be the vertex of $\C$ nearest to $\xhat$.
        \State Check if $v^0$ is the optimal solution to (\ref{proj:eq1}).
        \State If $v^0$ is not optimal, include new vertices, $v^r$ and search over corrals of increasing dimension.
    \end{algorithmic}
    \end{algorithm}

As we will see later for \SE and \BE, the algorithms designed with this approach as the basis are a lot more efficient for our problem solution than off-the-shelf algorithms that are known to be {\it generally efficient} for finding the projection \cite{duchi2008efficient}. Also, for the \Ass and \Slate cases, it is important to mention that no efficient projection algorithms exist in the literature. The algorithms that we develop for them here are new and are well-suited for our problem solution.

Before we give our algorithms for the specially structured polytopes in \ref{sec:structured}, let us discuss a useful general purpose, vertex-oriented projection algorithm for {\it General Polytope}. Many classical algorithms exist for solving the projection-to-convex-hull problem:
\begin{equation}
    \min_{x\in \C} \| x - \xhat \|^2 \mbox{ where } \C = co \, \{ v : v\in V \}
    \label{proj:eq2}
\end{equation}
The algorithms in~\cite{Gilbert1966, BG1969, Meyer1970, MDM1974, GJK1988, wolfe1976, chakrabarty2014provable} are a few of them. DuaLip can be made to work well with any one of these algorithms. More than the algorithm choice, the following three key pieces, which are all our contributions, are absolutely essential for efficiency.\footnote{For example, our initial implementation of using the Wolfe algorithm~\cite{wolfe1976, chakrabarty2014provable} without optimizations (a) through (c) made the projection 100x slower on a test problem.} (a) While these algorithms allow an arbitrary corral for initialization, our vertex-oriented approach always starts with the nearest-vertex single-point corral. (b) Even outside of these algorithms we check if this single-point corral is already optimal for (\ref{proj:eq2}) and, if so, these algorithms are never called. Note that the computations in (a) and (b) are an integral part of our Algorithm~\ref{alg:ourapp} (steps 1 and 2 there). (c) While these algorithms are set up for the projected point being the origin, doing a transformation of the convex hull points to meet this requirement would lead to a large inefficiency. It is important that we keep the original vertices as they are and carefully modify all operations of the algorithms efficiently. The summary of this is that, for {\it General Polytope}, DuaLip is not wedded specifically to any particular general-purpose algorithm but rather to the problem (\ref{proj:eq2}), and, for the class of problems and applications covered in the paper, we do a lot more to make a chosen algorithm to work efficiently than that algorithm itself.


\subsection{Wolfe's algorithm for General Polytope}
Due to its popularity, we delve into the classic algorithm given by Wolfe~\cite{wolfe1976, chakrabarty2014provable} for (\ref{proj:eq1}), which also nicely instantiates the approach in Algorithm~\ref{alg:ourapp} for the \GP case. We briefly outline this algorithm here. 
At a general iteration of the Wolfe algorithm, there is a corral $C$ formed using a subset of vertices $U$, and $x$, the projection of $\xhat$ to $C$.
Using optimality conditions it is easy to show that $x$ solves (\ref{proj:eq2}) iff 
\begin{equation}
    \min \{ (x-\xhat)^T v : v\in V\backslash U \} \ge (x-\xhat)^T x \label{proj:eq3}
\end{equation}
If (\ref{proj:eq3}) is violated then the inclusion of $v=\arg\min \{ (x-\xhat)^T v : v\in V\backslash U\}$ to the vertex set of $C$ to form the new corral, $\Ctil$ is guaranteed to be such that $\xtil$, the projection on $\Ctil$, satisfies $\|\xtil-\xhat\|^2 < \|x-\xhat\|^2$. 
Since it is a descent algorithm and the number of corrals are finite, the algorithm terminates finitely. For more details see the full paper \cite{dualip}. 


Wolfe's algorithm can be initialized with an arbitrary corral and $x$ as the projection to that corral. However, to be in line with our approach in Algorithm~\ref{alg:ourapp}, we initialize $x$ to be the vertex in $V$ that is closest to $\xhat$. This step is also simple and efficient, requiring just a linear scan over the vertices. It is also useful to note that the immediately following check of optimality, (\ref{proj:eq3}) is also well in line with our approach in Algorithm~\ref{alg:ourapp}; the Wolfe algorithm terminates right there if the projection is a vertex solution, which we know to be the case for most $i$. 

\section{Special Structured Polytope Projections}
\label{sec:structured}
Wolfe's algorithm as detailed in Section \ref{sec:projection} is highly efficient when the corral dimension is small. It can be made even more efficient for problems with a special structure such as \textit{Simplex}, \textit{Box-Cut} and \textit{Parity Polytope}.

{\bf Simplex Projection:}
For \SE, $\C = co \, \{v_k\}_{k=1}^K$ where $v_k$ is the vector whose $k$-th component is equal to $1$ and all the other components are equal to $0$. \ecl employs the algorithm from ~\cite[Algorithm 1]{duchi2008efficient} for \SE, 
with complexity $O(K\log K)$. The algorithm also requires all components of $\xhat$ to be available; in the cases of \Ass and \textit{Slate}, 
this is very expensive.

The algorithm of~\cite{duchi2008efficient} is inefficient for our problem since in most situations, the nearest vertex of $\C$, which can be done in $O(K)$ effort, is expected to be the projection. We can make it more efficient guided by the approach in Algorithm~\ref{alg:ourapp} and specifically, the Wolfe algorithm outlined earlier. The steps are given in Algorithm~\ref{alg:duchimod}. Since $\|v_j-\xhat\|^2 = 1 + \|\xhat\|^2 - 2 \xhat_j$, the first step is equivalent to finding the index, ${\ktil}$ that has the largest $\xhat$ component, which forms the initializing corral. The remaining steps correspond to sequentially including the next best index (Step 3), checking if it does not lead to any further improvement (Steps 4 and 5), and if so, stopping with the projection.
Step 3 is worth pointing out in more detail. It is based on \eqref{proj:eq3}. Because all vertices of $\C$ are orthogonal to each other and $x$ is a linear combination of the vertices in $U$, $x^Tv=0 \; \forall v\in V\backslash U$. Thus, $\arg\min \{ (x-\xhat)^Tv : v\in V\backslash U \} = \arg\max \{ \xhat^Tv : v\in V\backslash U \}$, leading to Step 3 of Algorithm~\ref{alg:duchimod}.

\SI can be efficiently solved by first obtaining $x^{\mbox{box}}$, the box projection. If $\sum_k x_k^{\mbox{box}} \le 1$, then $x^{\mbox{box}}$ is also the \SI projection; else, the \SE projection is the the \SI projection.

In the normal case where all components of $\xhat$ are available, the incremental determination of the next best vertex in step 3 can be done using the max heapify algorithm~\cite{Cormen2001introduction}. If the algorithm stops in $q$ steps, then the overall complexity is $\mathcal{O}(K + q \log K)$. 
\cite{vandenberg08} gives a similar algorithm especially for \SE, but without the geometrical intuition and the power of its extension to structured outputs.


\def\chat{\hat{c}}
\textbf{\textit{Simplex-Eq} Projection for Structured Outputs:}
Consider any structured output setting in which, for each $i$ there is a parameter vector, $\xi_i$ and a possibly large set of configurations indexed by $k=1,\ldots,K$ such that: (a) for each given configuration $k$, $c_{ik}$ and $a_{ik}^{(j)}\; \forall j$ can be easily computed using $\xi_i$; and (b) the incremental determination of the {\it next best $k$} according to the value, $\chat_{ik} = c_{ik} + \sum_j \lambda_j a_{ik}^{(j)}$ is efficient. Using a \SE setup it is clear that Algorithm~\ref{alg:duchimod} efficiently solves the projection problem for this case. Note that the $\chat_{ik}$, $c_{ik}$, $a_{ik}^{(j)}$ do not need to be computed for all $k$. Instead of $O(K)$, complexity is usually polynomial in terms of the dimension of $\xi_i$.
\Ass and \Slate are special cases of such a setup, with each $k$ corresponding to a certain path on a bipartite graph and a DAG, respectively.

\setlength{\textfloatsep}{2pt}
\begin{algorithm}[!tb]
   \caption{Modified Duchi et al. algorithm for \SE}
   \label{alg:duchimod}
\begin{algorithmic}[1]
   \State $k_1 \gets \arg\max_{k\in [K]} \xhat_k$, and set: $\K=\{k_1\}$
   \Do
   \State $\ktil = \arg\max_{k\in [K]\backslash\K} \xhat_k$. \Comment{next best vertex}
   \State $\alpha = \xhat_{\ktil} - \frac{1}{|\K|+1} \left(\sum_{r\in \K\cup\{\ktil\}} \xhat_{r} -1 \right)$.
   \State Set $\K \leftarrow \K \cup \{\ktil\}$.
   \doWhile{$\alpha > 0$}
   \State{$\theta = \frac{1}{\K} \left(\sum_{r\in\K} \xhat_{r} \right)$}
   \State\Return $x_r^* = \xhat^r-\theta \; \forall r\in\K$, $x_r^* = 0 \; \forall r\not\in\K$
\end{algorithmic}
\end{algorithm}

{\it Assignment projection:} Let us return to the notations given in the definition of Multiple-slot \Ass in Section~\ref{subsec:applications}. 
Let $\xi_i = \{ c_{i\ell s}, \{ a_{i\ell s}^{(j)} \}_j  \}_{\ell , s}$.
Each index $k$ corresponds to one assignment, i.e., a set $\{(l_t, s_t)\}_{t=1}^{S_i}$, with $c_{ik} = \sum_{t=1}^{S_i} c_{i\ell_t s_t}$ and $a_{ik}^{(j)} = \sum_{t=1}^{S_i} a_{i\ell_t s_t}^{(j)}$. Thus, we have $\xhat_{ik} = c_{ik} + \sum_{j=1}^m \lambda_j a_{ik}^{(j)} = \sum_{t=1}^{S_i} \phi_{i\ell_t s_t}$ where $\phi_{i\ell s} = c_{i\ell s} + \sum_{j=1}^m \lambda_j a_{i\ell s}^{(j)}$. Thus, to pick ordered elements of $\{\xhat_k\}_k$, we just need to consider an assignment problem with cost defined by $\phi_{i\ell s}$. Step 1 of Algorithm~\ref{alg:duchimod} corresponds to picking the best assignment. Step 3 corresponds to incrementally choosing the next best assignments. Efficient polynomial time algorithms for implementing these steps are well known~\cite{bourgeois1971, kuhn1955, murty1968}.

{\it Slate projection:} The specialization of Algorithm \ref{alg:duchimod} for this case is similar to what we described above for \Ass. Because ranking of items needs to be obeyed and there are edge costs between consecutive items in a slate, dynamic programming can be used to give an efficient polynomial time Algorithm~\cite{keerthi2007} for Step 1. For incrementally finding of the next best slate (Step 3), the ideas in~\cite{Haubold2018} can be used to give efficient polynomial time algorithms. It is clear from these specializations that Algorithm~\ref{alg:duchimod} is powerful and can be potentially applied more generally to problems with other complex structures.

\def\vhat{\hat{v}}
{\bf Box-Cut Projection:} For \BE, it is easy to see that $\C = co \, \{v_r\}_{r=1}^{K \choose p}$ where each $v_r$ is a vector with $p$ components equal to $1$ and all the other components equal to $0$. Without enumeration of the vertices, it is easy to apply Wolfe's algorithm. 
First, for any given direction $\eta\in R^K$ let us discuss the determination of $\max_r \eta^T v_r$. This is equivalent to finding the top $p$ components of $\eta$ and setting them to $1$ with all remaining components set to $0$. Using a heap this can be done efficiently in $\mathcal{O}(K)$ time. Let $\vhat(\eta)$ denote the vertex of the Box-Cut polytope thus found. 

The initializing step of the Wolfe algorithm is the finding of the vertex $x$ nearest to $\xhat$. Since $\|v_r-\xhat\|^2 = p + \|\xhat\|^2 - 2 \xhat^T v_r$ and the first two terms do not depend on $v_r$, this step is equivalent to finding $\vhat(\xhat)$. This nearest vertex is set as the initializing corral. A general step of the Wolfe algorithm requires the checking of (\ref{proj:eq3}). Since $U$ is a corral and $x$ is the projection of $\xhat$ to the affine hull of $U$, $(x-\xhat)^T u = (x-\xhat)^T x \; \forall u\in U$. Thus (\ref{proj:eq3}) can also be written as
\begin{equation}
    \min \{ (x-\xhat)^T v : v\in V \} \ge (x-\xhat)^T x. \label{proj:eq3a}
\end{equation}
So, all that we need to do is to set $\eta = (\xhat-x)$ and obtain $\vhat(\eta)$ to check (\ref{proj:eq3a}) as $-\eta^T\vhat(\eta) \ge -\eta^T x$ to stop the Wolfe algorithm or, if that fails, use $\vhat(\eta)$ as the next vertex for proceeding with the Wolfe algorithm. Since, by arguments of sections~\ref{sec:projection}, the number of steps of the Wolfe algorithm will be just $1$ for most $i$'s, the cost of doing \BE projection per $i$ is just $\mathcal{O}(K)$. This way of doing projection is much more efficient for use in our problem than more general algorithms with $\mathcal{O}(K \log K)$ complexity given in the literature~\cite{DaiF06}. 

\BI can be done efficiently similar to the \SI projection by using the \Box and the \BE projection. 

{\bf Parity Polytope Projection:} ADMM based LP decoding of LDPC codes~\cite{wei2017iterative} requires projections to a special polytope called the Parity (aka Check) polytope, which is $\C = co \, \{v_r\}$ where the $v_r$ are binary vectors with an even number of 1s. The projection algorithms in~\cite{lin2019, wei2017iterative} are specially designed for the case, but they are general purpose algorithms that use the inequalities-based representation of $\Ccal_i$ and are not designed to be efficient for the vertex oriented structure. 

The specialization of the Wolfe algorithm, that provides such an efficiency, is similar to that for \BE. The key computations are: (a) finding the vertex $v$ nearest to $\xhat$; and (b) checking (\ref{proj:eq3a}), which requires, for any given any direction $\eta$, $\vhat(\eta) = \arg\max_{v\in V} \eta^T v$. Let us start with (b). $\vhat(\eta)$ is a binary vector and we need to find the elements of $\vhat(\eta)$ having the 1s. Since we are maximizing $\eta^T v$, take the positive indices, $P=\{i: \eta_i >0 \}$. If $|P|$ is even, choose the set of elements of $\vhat(\eta)$ with 1s as $P$. If $|P|$ is odd, let $i_1$ be the index of the element of $P$ with the smallest $\eta_i$, and $i_2$ be the index of the element of the complement of $P$ with the largest $\eta_i$. If $\eta_{i_1} + \eta_{i_2} > 0$, include $i_1$ and $i_2$ in $P$; else, remove $i_1$ from $P$. This is the optimal way of choosing $|P|$ to be even. Then choose the set of elements of $\vhat(\eta)$ with 1s as $P$. Thus, the determination of $\vhat(\eta)$ requires just a linear scan of $\{\eta_i\}$.

Let us now come to the finding of the vertex $v$ nearest to $\xhat$, i.e., $\arg\min_v \|\xhat-v\|^2$. Now, $\|\xhat - v\|^2 = \|\xhat\|^2 + \|v\|^2 - 2\xhat^T v$. Since $\|\xhat\|^2$ is a constant in the finding of $v$, and, for binary vectors, $\|v\|^2 = e^Tv$ where $e$ is a vector of all 1s, we get $\arg\min_v \|\xhat-v\|^2 = \arg\max_v \eta^T v$, where $\eta=\xhat - 0.5e$. With $\eta$ set this way, we get $\arg\min_v \|\xhat-v\|^2$ to be $\vhat(\eta)$, whose determination was described in the previous paragraph. Thus, the finding of the vertex nearest to $\xhat$ also requires just a linear scan of $\{\eta_i\}$. It is worth trying the Wolfe algorithm as an alternative to existing algorithms to project points to parity polytopes in the ADMM based LP decoding of LDPC codes~\cite{wei2017iterative}.

\section{Experiments}
\label{sec:experiments}
\def\gbest{ g_0^{\mbox{best}} }
We demonstrate the value of the algorithms designed in \S\ref{sec:projection} and \S\ref{sec:structured} empirically using dataset $D$. We solve the problem, using different number of cluster nodes ($w$) and report the speedup in Table \ref{tab:projection} by comparing with the large scale \ecl solver. Note that, since \ecl only supported the \SE projection via off-the-shelf projection algorithms \cite{duchi2008efficient}, we focus our results on the \SE use case on their data \cite{basu2020} to showcase an apples to apples comparison of the overall LP solver.   
\begin{table}[ht]
    \centering
    \begin{tabular}{|l|c|c|c|c|} 
\hline
 \# nodes ($w$) & 55 & 85 & 165 & 800 \\
 $\ceil{I/w}$ (in $100k$) & 15	& 10 & 5 & 1 \\
 Speedup & 6x & 6.5x & 7.5x & 8x \\
\hline
\end{tabular}
    \caption{\small{ Speedup is the ratio of time taken by the DuaLiP projection algorithm to that of \ecl (based on ~\cite{duchi2008efficient}). 
    }}
\label{tab:projection}
\end{table}
The speedup measures the improvement in time taken to aggregate $I/w$ projection operations. Note that for large values $I/w$, both methods have a fixed cost of going over a lot of data instances that reduces the relative speedup of the DuaLip projection. From Table~\ref{tab:projection}, it is clear that even for a simple projection type such as \SE, employing off-the-shelf algorithms \cite{duchi2008efficient} may not be ideal. Basu et. al.\cite{basu2020} show that commercial solvers (SDPT3, CPLEX, CLP, GUROBI) do not scale to the problem-sizes in web-scale marketplaces and use a distributed version of the Splitting Conic Solver (\scs\cite{scs}) to establish a baseline for scale. We benchmark \dlip against \ecl and \scs in Table~\ref{tab:scale-results} on real data to show a $7$x improvement over the state-of-the-art. \ecl and \dlip exploit the data sparsity in the constraint matrix ($A$) often present in marketplace data unlike \scs and other commercial solvers. Additionally, \dlip benefits from efficient projection (\S\ref{sec:projection} and \S\ref{sec:structured}). We observe similar large efficiency gains on multiple applications of comparable size using our incremental projection algorithm. Tables~\ref{tab:projection} and \ref{tab:scale-results} show that the speed-up depends less on the dataset and is more a function of the projection and $I/w$.

\begin{table}[!ht]
    \centering
    \resizebox{0.9\columnwidth}{!}{%
    \begin{tabular}{|l|c|c|c|c|} 
\hline
\multirow{2}{*}{Problem} & \multirow{2}{*}{Scale $n$} & \multicolumn{3}{c|}{Time (hr)}\\
 && \dlip & \ecl & \scs \\
\hline
\multirow{3}{10em}{Email Optimization \cite{basu2020} $nnz(A) = 10I \approx 1B$} & $10^7$ & 0.11 & 0.8 & 2.0\\
& $10^8$ & 0.16  & 1.3 & >24 \\ 
& $10^9$ & 0.45 & 4.0 & >24 \\ 
\hline
\multirow{3}{10em}{Item Matching \eqref{eq:marketplace} $nnz(A) = 100 I \approx 10B $} & $10^{10}$ & 0.62 & 4.5 & >24\\
& $10^{11}$ & 1.1 & 7.2 & >24  \\ 
& $10^{12}$ & 2.60 & 11.9 & >24 \\ 
\hline
    \end{tabular}
}
    \caption{\small{Run time (hrs) for extreme-scale problems on real data. Here, $nnz(A)$ denotes the number of non-zero entries in $A$ and all runs are on Spark $2.3$ clusters with up to $800$ processors.}}
    \label{tab:scale-results}
\vspace{-0.4cm}
\end{table}

\section{Discussion}
\label{sec:conclusion}
Through empirical evidence and guided by theory, we have developed novel algorithms for efficient polytope projections under a vertex oriented structure. Such structures commonly occur in very large-scale industrial applications. Using these novel algorithms we were able to achieve a drastic overall speedup compared to the state-of-the-art while solving such web-scale Linear Programs. This scalability allows us to expand to a much larger class of problems with complex structured constraints such as \Slate and \Ass. It also opens up the applicability in semi-supervised learning, structured prediction on text data, optimal marketplace auctions~\cite{edelman2007internet}, and rank order of items. All trends reported in this paper hold across many web marketplace problems (\S\ref{subsec:applications}). We provide results from one large-scale internal dataset ($D$) for clarity of impact and open-sourced the solver on Github\footnote{https://github.com/linkedin/dualip} with benchmarks on a public-domain dataset for reproducibility. 





\bibliography{paper}
\bibliographystyle{abbrv}

\clearpage
\appendix

\section{Proofs}
\label{app:projection}

This section contains various details related to \S\ref{sec:projection} and \S\ref{sec:structured} of the paper.

A result is said to hold in general position if the result holds almost surely in the space of perturbations and other parameters; in this section, these are $\delta c$, $\delta b$, $\delta d$, $\delta e$, $\delta p$, $\lambda$ and $z$. We will simply use the phrase `general position' and it will be clear from the context as to which perturbations and parameters it means.

\subsection{Proof of part 1 of Theorem 1}
\label{subsec:appprojth11}

\def\P{{\cal{P}}}
\def\D{{\cal{D}}}
\def\O{{\cal{O}}}
\def\nact{n_{act}}

The proof is based on the following lemma, which we establish later. 


{\bf Lemma B.1} Consider the LP
\begin{equation}
    \min_x~~ (c + \delta c)^T x ~ \mbox{ s.t. } ~ Px \le p + \delta p, ~ Ex = e + \delta e
    \label{appprojth1:eq1}
\end{equation}
where the feasible set is a compact subset of $R^n$ and $\P$ is any open set of allowed perturbation values $(\delta c, \delta p, \delta e)$. Then, 
in general position
the solution $x$ of (\ref{appprojth1:eq1}) is a unique point at which exactly $n$ constraints of $Px\le p+\delta p$, $Ex = e + \delta e$ are active. $\qedsymbol$

We begin with a clarification concerning the mean optimal corral dimension associated with $x^* = \arg\min_{x}~c^T x ~\mbox{s.t.}~Ax \leq b,~x_i \in \C_i, i \in [I]$. Though $x^*$ does not originate from a projection operation, it is actually a projection. This comes from Lemma 2 of \cite{basu2020}, which implies that there exists $\bar{\gamma}>0$ such that for all $0 < \gamma \le \bar{\gamma}$, $x^*$ is the projection associated with the determination of $g_\gamma(\lambda_\gamma)$ where $\lambda_\gamma = \arg\max_\lambda g_\gamma(\lambda)$. Thus, it is sensible to talk of the mean optimal corral dimension, $\mu(x^*)$.

To prove part 1 of Theorem 1, let us apply Lemma B.1 to the case where we combine $Ax\le b+\delta b$ and $Dx\le d+\delta d$ to form $Px\le p + \delta p$. The equality constraint, $Ex = e + \delta e$ stays as it is. The lemma implies that 
in general position
the solution $x^*$ of the LP is unique and the number of active constraints is exactly equal to $n=IK$. Among the active constraints, some of them, call this number $m_{act}$, come from $Ax\le b$. The rest of the active constraints come from $Dx\le d$, and $Ex = e + \delta e$ i.e., the $\C_i$, which, in terms of optimal corral dimensions, is equal to\footnote{The number of active constraints is related to the representation of $\C_i$ by inequality constraints while optimal corral dimension is related to the representation using vertices. We are able to connect the two due to the special nature of the $\C_i$. For the {\it General Polytope} case we would need $Dx\le e + \delta e$ and and $Ex = e + \delta e$ to be a minimal representation of $\C$.} $\sum_i (K-\mu_i)$ for \SE {\it (Iq)}, {\it Assignment} and {\it Slots}, and upper bounded by $\sum_i (K-\mu_i)$ for \BE {\it(Iq)}. (It is easy to see that these structures are unchanged if the perturbation $(\delta d, \delta e)$ is small.) Thus,
\begin{equation}
    n \le m_{act} + \sum_i(K-\mu_i) = m_{act} + n - n\mu(x)
    \label{appprojth1:eq1a}
\end{equation}
Since $m_{act}\le m$, we get $\mu(x^*) \le m/n$.

\subsection{Proof of part 2 of Theorem 1}
\label{subsec:appprojth12}

Similar to Lemma B.1 we can prove the following lemma.

{\bf Lemma B.2} Consider the LP
\begin{equation}
    \min_x~~ (c + A^T\lambda + \delta c)^T x ~ \mbox{ s.t. } ~ Dx \le d + \delta d, ~ Ex = e + \delta e
    \label{appprojth1:eq1b}
\end{equation}
where the feasible set is a compact subset of $R^n$ and $\D$ is any open set of allowed perturbation/parameter values $(\delta c, \delta d, \delta e, \lambda)$. Then, 
in general position
the solution $x$ of (\ref{appprojth1:eq1b}) is a unique point at which exactly $n$ constraints of $Dx\le d+\delta d$, $Ex = e + \delta e$ are active. $\qedsymbol$

To prove part 2 of Theorem 1 we just apply Lemma B.2. Since $Ax\le b$ constraint is absent, analysis similar to the proof of part 1 yields that 
in general position
we have $\mu(x_0^*(\lambda))=0$.

\subsection{Proofs of Lemmas B.1 and B.2}
\label{subsec:appprojlB1}

The two lemmas are geometrically intuitive; however, proving them rigorously is somewhat technical and detailed.

Let us start by proving Lemma B.1. The Lagrangian for the LP in (\ref{appprojth1:eq1}) with dual vectors $\psi$ and $\phi$ is
\begin{equation}
    L = (c + \delta c)^T x + \psi^T(Px - p - \delta p) + \phi^T(Ex - e - \delta e)
    \label{appprojth1:eq2}
\end{equation}
A triple $(x,\psi,\phi)$ is optimal iff the following KKT conditions hold:
\begin{eqnarray}
    P^T\psi + E^T\phi + c + \delta c = 0, ~ Px \le p + \delta p, \label{appprojth1:eq3} \\
    \psi\ge 0, ~ \psi^T(Px - p - \delta p) = 0, ~ Ex = e + \delta e
    \label{appprojth1:eq4}
\end{eqnarray}
Let
\begin{equation}
    S_1 = \{ r : (Px-p-\delta p)_r = 0 \}, ~ S_2 = \{ r \in S_1 : \psi_r = 0 \}
    \label{appprojth1:eq5}
\end{equation}
$S_1$ is the set of active (primal) inequality constraints and $S_2\subset S_1$ is the subset with positive $\psi$ values. Note that not all the $\psi_r, r\in S_1$ are required to be positive. Note also that $\psi_r=0 \; \forall r\not\in S_1$. 
Let us rewrite the KKT conditions in (\ref{appprojth1:eq3}) and (\ref{appprojth1:eq4}) as
\begin{eqnarray}
    P^T\psi + E^T\phi + c + \delta c = 0, ~ (Px - p - \delta p)_{S_1} = 0, \label{appprojth1:eq6} \\
    (Px - p - \delta p)_{S_1^c} < 0, ~  
    (\psi)_{S_1\backslash S_2} > 0, ~ (\psi)_{S_1^c\cup S_2} = 0
    \label{appprojth1:eq7} \\
    Ex -e - \delta e = 0 \label{appprojth1:eq7a}
\end{eqnarray}
In the above, we are analyzing what happens at a solution $(x,\psi,\phi)$. Let us turn this around a bit and look at the (finite) collection of all $(S_1, S_2)$ possibilities satisfying $S_2\subset S_1$, and, for a given $(S_1, S_2)$ look at the set of all optimal solutions $(x,\psi,\phi)$ satisfying (\ref{appprojth1:eq6})-(\ref{appprojth1:eq7a}), call it $\O (S_1,S_2)$. Note that $\O(S_1,S_2)$ will turn out to be empty for most $(S_1,S_2)$ pairs, which is fine. Using this definition, the set of all possible solutions of the LP in (\ref{appprojth1:eq1}) can be written as
\begin{equation}
    \O = \cup_{(S_1,S_2)} \O(S_1,S_2) \label{appprojth1:eq8}
\end{equation}

Let us take one $(S_1, S_2)$. We will use Transversality theorem (a corollary of Sard's theorem) to show that $\O(S_1,S_2)$ is at most a singleton 
in general position.
We state Transversality theorem~\cite{guillemin} in the simplest form that suffices our purpose.

\def\Ytil{\tilde{Y}}

{\bf Transversality theorem} Suppose $F: Y\times Z \to R^N$ is a smooth differentiable function such that (a) $Y$ and $Z$ are open sets in $R^L$ and $R^M$ and (b) the Jacobian of $F$ with respect to $(y,z)$ has full row rank for all $(y,z)\in Y\times Z$. Then 
in general position
the set $\Ytil(z) = \{ y\in Y : F(y,z) = 0 \}$ is either empty or a differentiable manifold of dimension $(L-N)$. $\qedsymbol$

To set up Transversality theorem for proving Lemma B.1, let $y=(x,\psi,\phi)$, $z=(\delta c, \delta p, \delta e)$, $Z=\P$,
\begin{eqnarray}
 Y = \{(x,\psi,\phi) : (Px - p - \delta p)_{S_1^c} < 0, ~ (\psi)_{S_1\backslash S_2} > 0 \} \label{appprojth1:eq9}
\end{eqnarray} 
and the function, $F:Y\times Z \to R^N$ as
\begin{equation}
 F = \left( 
 \begin{array}{c}
    P^T\psi + E^T\phi + c + \delta c \\
    (Px-p-\delta p)_{S_1} \\
    Ex - e - \delta e \\
    (\psi)_{S_1^c\cup S_2}
 \end{array}
 \right) \label{appprojth1:eq9a}
\end{equation}
so that $\O(S_1,S_2)$ can be written as
\begin{equation}
 \O(S_1,S_2) = \Ytil(z) = \{ y \in Y : F(y,z) = 0 \} \label{appprojth1:eq11}
\end{equation}
Note that $Y$ and $Z$ are open sets.
If we look at the Jacobian of $F$ with respect to the subset of variables, $(\delta c, (\delta p)_{S_1}, \delta e, (\psi)_{S_1^c\cup S_2})$ it is a non-singular diagonal matrix with $1$'s and $-1$'s in the diagonal. Hence the Jacobian of $F$ with respect to $(y,z)$ has full row rank. Thus, by Transversality theorem, 
in general position
the set $\O(S_1,S_2)$ is either empty or a differentiable manifold of dimension $(L-N)$. With $n$ being the number of variables in $x$ and $\ell$ being the number of constraints in $Px\le p + \delta p$, $Ex\le e + \delta e$ note that $L = n + \ell, ~ N = n+ \ell + |S_2|$.
Let's consider two cases.

{\bf Case 1.} $\mathbf{|S_2|> 0.}$ Since $(L-N)$ is negative, it automatically means that 
in general position
the set $\O(S_1,S_2)$ is empty. Thus, in general position, we don't have to consider the case $S_2\not= \emptyset$.

{\bf Case 2.} $\mathbf{|S_2|= 0.}$ For this case, $(L-N)=0$ and, 
in general position
the set $\O(S_1,S_2)$ is a zero dimensional manifold, i.e., a set of isolated points. By convexity of $\O(S_1,S_2)$, it can be at most a singleton.  Also, if $\O(S_1,S_2)\not=\emptyset$, then, since $(Px-p-\delta p)_{S_1} = 0$ and $Ex - e - \delta e = 0$ independently determine $x$, and, $P^T\psi + c + \delta c = 0$, $(\psi)_{S_1^c}=0$ together determine $\psi$, we must have $|S_1|=n$; otherwise, in general position, one of the two equation sets will have no solution.

Let us now come to $\O$ in (\ref{appprojth1:eq8}). 
In general position,
convexity of $\O$ implies that the only possibility is that there is a unique $S_1$ such that $\O = O(S_1,\emptyset)=\{(x^*,\psi^*,\phi^*)\}$, the unique optimal solution of (\ref{appprojth1:eq1}). $\qedsymbol$

The lines of proof for Lemma B.2 are similar. $Px\le p + \delta p$ is replaced by $Dx\le e + \delta e$ and there is an additional parameter $\lambda$ that is included in $z$.

\subsection{Wolfe's algorithm}
\label{subsec:appprojwolfe}

In \S\ref{sec:projection} we described the essence of Wolfe's algorithm. Here we give fuller details in Algorithm~\ref{alg:wolfe}. 

Steps 5-22 form one iteration of the algorithm going from one corral to a better corral. Steps 5 and 6 correspond to checking optimality of $x$, the projection to the current best corral; these steps are same as each of (\ref{proj:eq3a}) and (\ref{proj:eq3}). Step 5 forms the most expensive part of the algorithm. Note that, in problems with structure in the vertices of $\C$, step 5 can be done very efficiently without enumerating all vertices. As we explained in \S\ref{sec:structured}, this is exactly what happens in \Ass, \Slate and \BE {\it (Iq)}.

If checking of optimality of $x$ fails, the algorithm enters the minor cycle in which a next, better corral is identified efficiently. Steps 11-22 form the minor cycle. Step 12 requires the projection of $\xhat$ to $\tt{aff}(S)$, the affine hull of the vertices in $S$, which can be obtained by solving a system of linear equations in $(|S|+1)$ unknowns. Since the corral dimensions $(|S|-1)$ encountered in marketplace LPs are small, the affine hull computation is cheap.

In many cases of $\C$, the minor cycle breaks at the first cycle/entrance to step 14. The modified Duchi et al algorithm for \SE (algorithm \ref{alg:duchimod}) is an example. This is also true in most iterations of other cases of $\C$. 

\begin{algorithm}[!ht]
\caption{Wolfe's algorithm}
\label{alg:wolfe}
\begin{algorithmic}[1]
  \Statex {\textbf{Input: } $\xhat$, $\mathcal{B} =$ Vertex set of $\C$ }
  \Statex {\textbf{Output: } $x^* = \arg\min_{x\in\C} \|x-\xhat\|^2$ }
  \State{In all iterations we maintain $S=\{v_r\}$, a subset of vertices, and $x = \sum_{r \in S} \rho_r v_r$ as a convex combination of the vertices in $S$. At the end $S$ forms the optimal corral. }
  \State{Find $v$ the vertex closest to $\xhat$.}
  \State{Initialize $x \gets v$, $S \gets \{v\}$ and $\rho_1 \gets 1$. }
  \While{true}\Comment{Major Cycle}
    \State{$v \mathrel{:}=\arg \min_{p \in \mathcal{B}} (x-\xhat)^Tp$ }
    \If{$(x-\xhat)^Tx \le (x-\xhat)^Tv$}
        \State{Set $x^*=x$.}
        \State{\textbf{break}}
    \EndIf
    \State{$S \mathrel{:}= S \cup \{v\}$, $\rho_{|S|} \gets 0$ }
    
    \While{true}\Comment{Minor Cycle}
        \State{Find $y = \arg \min_{z\in \tt{aff}(S)}\|z-\xhat\|^2$ and} \State{$\alpha$ s.t. $y = \sum_{r\in S}\alpha_r v_r$}
        \If{$\alpha_r \ge 0 \; \forall_r$}
            \State{\textbf{break}}\Comment{$y \in \tt{co}(S)$, so end minor cycle}
        \Else\Comment{If $y \not \in  \tt{conv}(S)$, then update $x$ to the intersection of the boundary of $\tt{co}(S)$ and the segment joining $y$ and previous $x$. Delete points from $S$ which are not required to describe the new $x$ as a convex combination.}
        \State{$\theta \mathrel{:}= \min_{r:\alpha < 0} \frac{\rho_r}{\rho_r - \alpha_r}$}\Comment{Since $x=\sum_r\rho_rv_r$}
        \State{$x \gets \theta y + (1-\theta) x$}\Comment{Using $\theta$, the new $x$ lies in $\tt{conv}(S)$.}
        \State{$\rho_r \gets \theta \alpha_r + (1-\theta) \rho_r \; \forall r$}\Comment{Sets the coefficients of the new $x$}
        \State{$S \gets \{r: \rho_r > 0\}$}\Comment{Delete points which have $\rho_r = 0$. This deletes at least one point.}
        \EndIf
    \EndWhile
    \Comment{After the minor loop terminates, $x$ is updated to be the affine minimizer of the current set $S$}
    $x \gets y$
  \EndWhile
\State {\textbf{return }$x^*$}
\end{algorithmic}
\end{algorithm}


\end{document}